%% file: main.tex
\documentclass[10pt,twocolumn,letterpaper]{article}

\usepackage{cvpr}
\usepackage[accsupp]{axessibility} 
\usepackage{times}
\usepackage{epsfig}
\usepackage{graphicx}
\usepackage{amsmath}
\usepackage{amssymb}
\usepackage{enumitem}
\setlist{nolistsep,leftmargin=*}
\usepackage{amsfonts}
\usepackage{booktabs}
\usepackage{subcaption}
\usepackage{array,multirow}
\usepackage[pagebackref=true,breaklinks=true,colorlinks,bookmarks=false]{hyperref}

\makeatletter
\renewcommand{\paragraph}{%
  \@startsection{paragraph}{4}%
  {\z@}{0.5ex \@plus 0.5ex \@minus .2ex}{-1em}%
  {\normalfont\normalsize\bfseries}%
}
\makeatother

\begin{document}

\title{Human Body Model based ID using Shape and Pose Parameters%
\stepcounter{footnote}%
}


\author{
Aravind Sundaresan\thanks{Corresponding author: \href{mailto:aravind.sundaresan@sri.com}{aravind.sundaresan@sri.com}} \quad 
Brian Burns \quad 
Indranil Sur \quad \\
Yi Yao \quad
Xiao Lin \quad 
Sujeong Kim \\
\vspace{-.5em}\\
SRI International, 
333 Ravenswood Ave, Menlo Park, CA 94025, USA
}

\maketitle
\thispagestyle{empty}
\input{abstract}
\input{hmid}

\section*{Acknowledgments}
This research is based upon work supported in part by the Office of the Director of National Intelligence (ODNI), Intelligence Advanced Research Projects Activity (IARPA), via 2022-21110800003. The views and conclusions contained herein are those of the authors and should not be interpreted as necessarily representing the official policies, either expressed or implied, of ODNI, IARPA, or the U.S. Government. The U.S. Government is authorized to reproduce and distribute reprints for governmental purposes notwithstanding any copyright annotation therein.

\clearpage

{\small
\bibliographystyle{ieee}
\bibliography{hmid}
}
\end{document}

%% file: abstract.tex
\begin{abstract}
    We present a Human Body model based IDentification system (HMID) system that is jointly trained for shape, pose and biometric identification. HMID is based on the Human Mesh Recovery (HMR) network and we propose additional losses to improve and stabilize shape estimation and biometric identification while maintaining the pose and shape output. We show that when our HMID network is trained using additional shape and pose losses, it shows a significant improvement in biometric identification performance when compared to an identical model that does not use such losses. The HMID model uses raw images instead of silhouettes and is able to perform robust recognition on images collected at range and altitude as many anthropometric properties are reasonably invariant to clothing, view and range. We show results on the USF dataset as well as the BRIAR dataset which includes probes with both clothing and view changes. Our approach (using body model losses) shows a significant improvement in Rank20 accuracy and True Accuracy Rate on the BRIAR evaluation dataset.
\end{abstract}

%% file: hmid.tex
\section{Introduction}
\label{sec:introduction}

Biometric recognition technology has advanced significantly in recent years. However, it has primarily been developed using datasets that do not fully address the challenges of outdoor sequences, lower resolution, longer ranges and elevated view points common in advanced security, forensic and military applications. Cornett et al.~\cite{cornett2023expanding} have created the BRIAR dataset to bridge this data gap. In this paper, we present a novel biometric system designed to address the algorithmic challenges of the BRIAR dataset. Our Human body Model based biometric IDentification (HMID) system estimates biometric features based on human 3D body pose and shape. It exploits (but is not limited to) features tied to the anthropometric properties of the subjects for discrimination. Given that these features are a function of the whole body (with its much larger pixel area than just the face) and are based on the 3D aspects of the body, we believe that HMID is a robust approach to the challenges of long-range data, low image quality, elevated view points and view invariance.  In addition, we augment the training of HMID by using degraded images and forcing its shape output to reproduce non-degraded shape data in the form of silhouettes extracted from the original, high-resolution versions of the images. Thus, unlike other approaches, we do not depend on extracting silhouettes from degraded images during testing and deployment, but instead only depend on them from non-degraded images for our training shape losses.

\begin{figure*}[htb]
    \centerline{\includegraphics[width=\linewidth]
    {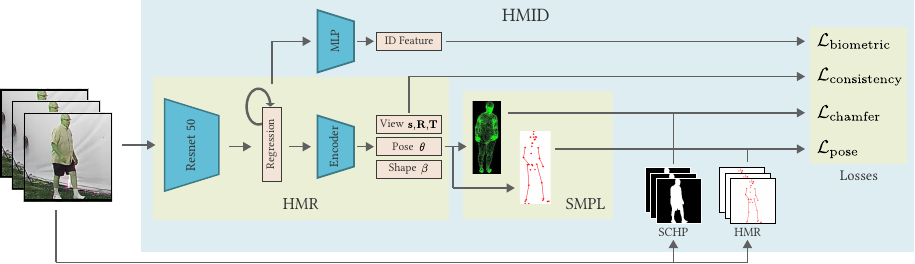}
    }
    \caption{Our Human body Model based biometric IDentification (HMID) system estimates biometric features straight from RGB image sequences. During training (shown), the inputs include an RGB sequence, corresponding silhouettes, and estimated ``ground truth'' 2D keypoints. Also during training, the outputs include the biometric features produced by the HMID model, as well as estimated 3D shape parameters, 2D keypoints and projected surface mesh vertices. Losses representing expectations for all these outputs drive the HMID to improve its internal representation and the resulting biometrics. The subjects in the BRS dataset have given permission for the images to be used in publications. }
    \label{fig:overview}
\end{figure*}

At its core, the algorithm uses the Human Mesh Recovery (HMR) model~\cite{kanazawa2018_hmr}. In addition to the Skinned Multi-Person Linear (SMPL) human model~\cite{loper2015smpl} related outputs (2D and 3D keypoints, shape and pose parameters, camera view, mesh vertices), we learn an additional biometric identification feature vector using a combined neural architecture. In~\cite{li2020_end-to-end}, HMR is also included in a whole-body biometric system. Our system differs with theirs in the features extracted from the HMR deep network and in the loss functions used to train this network. We believe that our features and losses make our system more robust with respect to challenges such as outdoor imagery that are not studied in~\cite{li2020_end-to-end}. To ensure robust matching, features from multiple frames and video samples of the same subject are used. Additionally, we explore aggregating features from input sequences based on the yaw and pitch angle of the view per frame. We have trained our model on the BRIAR Research Set (BRS) data~\cite{cornett2023expanding}\footnote{The subjects in the BRIAR datasets have consented to the public use of their imagery.} and show our results on the BRIAR Test Set (BTS) data and, in a cross-collection study, on the USF dataset~\cite{sarkar2005_hid}. We have have chosen to work with these collections because they emphasize outdoor data and the complications of varied backgrounds, clothing, distance, and view (including pitch). We compare our system with other works that have performed on this data (\S\ref{sec:exp:briar} and \ref{sec:exp:usf}). See \cite{cornett2023expanding} and \cite{sarkar2005_hid} for more on the objectives, adequacy and details of these collections.

HMID is an end-to-end model that starts with the front-end of the Human Mesh Recovery (HMR) model, which produces a raw 2048D feature vector using a Resnet50 DNN model as shown in Figure~\ref{fig:overview}. From this vector, via the HMR encoder, HMID estimates the camera view, SMPL model  shape and pose parameters. Also from this raw vector, HMID computes an ID feature vector using a linear layer (other feature vectors are also considered). HMR was trained using an extensive collection of body pose-annotated data (i.e., joint positions), but less so on shape-annotated data (e.g., body proportions). Since shape is an important aspect of whole-body ID, it is critical to improve it. To this end, during training, we not only train the ID features on biometric losses but also continue refining the HMR shape estimates on new data, while constraining the pose estimates from deviating too much from the original pre-trained HMR pose estimates using additional shape and pose losses. HMID estimates the projected surface mesh vertices (a silhouette approximation) and the projected human body model 2D key points from the SMPL model shape and pose parameters. During training, these outputs are compared to subject silhouettes extracted from the training videos using the method in \cite{li2020_schp} and 2D human key points estimated using the pretrained HMR model. As mentioned earlier, this training step also uses degraded input images to make HMID more robust with respect to range and resolution. \S\ref{sec:exp:briar} compares the biometric performance of HMID trained using our losses against HMR without this fine-tuning step.
Our work makes the following contributions:
\begin{enumerate}
    \item A 3D model-based approach that is end-to-end and computes biometric features directly from the penultimate, raw HMR feature layer. \S\ref{sec:related_work} discusses these features and \ref{sec:exp:ablation} experimentally compares them with other strategies, such as the one used in~\cite{li2020_end-to-end}.
    \item An additional Chamfer distance-based silhouette matching loss during training that clearly improves the HMR estimates of body shape and produces a promising whole-body biometric feature vector. \S\ref{sec:training:losses} discusses and compares loss strategies.
    \item A demonstration of the effects of different architecture variants, training losses and multi-sample aggregation methods on biometric performance (see \S\ref{sec:exp:ablation}).
    \item A cross-collection study showing promising domain robustness, in which one collection is used to train the HMID biometrics and a different collection is used to evaluate it by comparing with other methods (see \S\ref{sec:exp:usf}).
\end{enumerate}

\section{Related work}
\label{sec:related_work}

There is an extensive body of work concerned with the computation of biometric features associated with body shape and motion from one or more image sequences of a subject~\cite{nalty2022brief}. Several challenges need to be met for a generally applicable whole-body based biometric system: varying views (including changing pitch angles), reduced image quality from viewing at range, variable clothing and partial occlusion of the subject.
The core of our proposed approach is end-to-end dynamic model-based recognition using the extensively pre-trained and available human mesh recovery (HMR) network~\cite{kanazawa2018_hmr}, which estimates human 3D shape and pose parameters. We combine HMR with additional layers for computing biometric features from the raw, pre-pose HMR feature layer, as well as a temporal integration layer, all trained in an end-to-end fashion. This design is in contrast with the most common, appearance-based approaches to gait recognition that usually require extracted foreground (body) mask input and compute the biometric features using typically generic deep neural architectures and basic person identification losses~\cite{han2005individual,shiraga2016geinet,wu2016comprehensive,takemura2017input,wang2011human}. Typically in these methods, nothing besides training with biometric losses is specifically done to make the features less sensitive to cross-view matching, which remains a challenge. One recent appearance-based approach~\cite{myers2023recognizing} varies the process by using the image directly as input (not the silhouette) and, before biometric training, pre-trains the network to predict thirty linguistically-derived body descriptors (e.g., ``broad shoulders'') using hand-annotated data. The biometric output of this network does improve performance when fused with the output of a non-pretrained network (see~\ref{sec:exp} for discussion on BRIAR data).

In contrast with appearance-based approaches are methods that use architectures and training that explicitly disentangles view from 3D human shape and pose. One variant of this approach is model-based, which is becoming more popular given the availability of body pose extraction code~\cite{wagg2004automated, yam2004automated, liao2017pose, liao2020model}. A two-step process is typically used in these methods where 2D or 3D body pose sequences are first extracted, and then biometric features are computed from them in a separate processing step. Though this type of method is in principle extracting pose information that is less view-sensitive, the performance so far has been inferior to the other methods and sensitive to poor image quality, a key challenge.
In contrast to the two-step model-based approach, we developed a solution that is inspired by a recently reported alternative: end-to-end model-based gait recognition~\cite{li2020_end-to-end,li2021_multiview}, which has been shown to be superior to the other methods on the CASIA-B and OU-MVLP cross-view gait recognition challenges. In CASIA-B, their overall Rank-1 results beat the others at 89.53$\%$ for 124 subjects when including all cross-view and cross-condition matching (carrying / not carrying, wearing / not wearing heavy coat). This is particularly impressive given that the high-contrast, silhouette-like data in these collections are geared to the alternative silhouette-based methods. With more general clothing, backgrounds and lighting, for which the HMR front end of~\cite{kanazawa2018_hmr} is trained, the other methods should fare even worse. This approach also permits end-to-end training, which allows us to more easily fine-tune the system to meet the key challenges of more complex data sets, such as BRIAR~\cite{cornett2023expanding}, which includes range-induced image degradation and high-pitch views. This fine-tuning is possible given our mix of shape, pose, motion and biometric training losses. Other benefits of an end-to-end differentiable body biometric module include faster processing than multi-step methods and joint training with other biometrics such as face recognition.

While the work in~\cite{li2020_end-to-end, li2021_multiview} shows promise, it depends on the extracted HMR end-products (e.g., body shape parameters) as biometric features, which are not as reliable and expressive as the raw 2048D, penultimate HMR feature layer when processed further into biometric features, as is done in our work. In addition, the shape refinement in~\cite{li2020_end-to-end,li2021_multiview} is based on a differentiable rendering error, which is not as strong a correction signal for improving the shape parameters as the Chamfer distance loss our method uses. This loss is a good approximation to point-to-point alignment errors induced by shape mismatches when the pose error is low, as is typical of HMR. Moreover, we show qualititative results of the SMPL model shape parameter estimation (Figure~\ref{fig:fine-tuning}) illustrating the effectiveness of our approach, which was missing in \cite{li2020_end-to-end}. 

While the basic temporal smoothing approach of~\cite{li2020_end-to-end} was effective on more typical datasets, we believe that a stronger motion model, accounting for full 3D body dynamics, will be required to compensate for the degraded, low-resolution image data and larger view range that occurs in collections like BRIAR. We believe that this could be achieved by adding a component to the end-to-end system that enforces more realistic, human-like joint correlations and accelerations in the results. Computing realistic 3D body motions from image sequences is currently an active area of research~\cite{kanazawa2019learning,kocabas2020_VIBE}. One method demonstrating superior results in video-based 3D pose extraction challenges is VIBE~\cite{kocabas2020_VIBE}. In VIBE, a component is added after the HMR front end that uses a recurrent (GRU) layer to provide temporal filtering before the final 3D pose extraction layers. The trained weights of the densely connected GRU essentially enforce a human-like temporal and cross-joint consistency in the 3D pose and shape output. Results reported in~\cite{kocabas2020_VIBE} on the challenging 3D pose estimation benchmarks 3DPW show a 36$\%$ reduction in 3D pose error when the GRU-based temporal integration is added compared to per-frame pose estimates without it and 28$\%$ less error than the best competing temporal integration method~\cite{kanazawa2019learning}. Something like the GRU integration layer can easily be added to our end-to-end model-based design, and it is our plan to do so in future versions of our system. 

\section{Model training}
\label{sec:training}

\begin{figure}
  \centering
  \includegraphics[width=\linewidth]{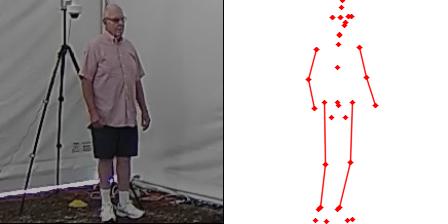} 
    \caption{Pose loss: The figures show the input image and the 2D keypoints estimated by the HMR model for 49 points. The keypoint loss estimated from the HMR model is compared against the ground truth 2D keypoints. }
  \label{fig:pose_loss}
\end{figure}

\begin{figure*}[tbh]
  \centering
  \includegraphics[width=\linewidth]{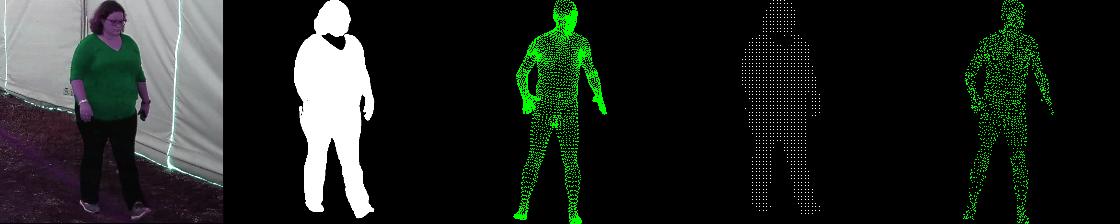} 
  \begin{subfigure}{0.195\linewidth}
      \caption{Input image}
      \label{chamfer:image}
  \end{subfigure}
  \begin{subfigure}{0.195\linewidth}
      \caption{Input Silhouette}
      \label{chamfer:silh}
  \end{subfigure}
  \begin{subfigure}{0.195\linewidth}
      \caption{Vertices}
      \label{chamfer:vert}
  \end{subfigure}
  \begin{subfigure}{0.195\linewidth}
      \caption{Silhouette cloud}
      \label{chamfer:silh_cloud}
  \end{subfigure}
  \begin{subfigure}{0.195\linewidth}
      \caption{Sampled cloud}
      \label{chamfer:vert_cloud}
  \end{subfigure}
    \caption{Chamfer shape loss: The figures above show the input image (\ref{chamfer:image}), input silhouette (\ref{chamfer:silh}), and the projected vertices from the output of the HMR model (\ref{chamfer:vert}). A 2D cloud is sampled from the silhouette image (\ref{chamfer:silh_cloud}). The 3D vertices are adaptively sampled to obtain a 2D vertices cloud (\ref{chamfer:vert_cloud}) so that the projection in the image is more uniformly distributed in space to avoid a bias in the Chamfer distance. The Chamfer distance is computed between the sampled silhouette (\ref{chamfer:silh_cloud}) and the adaptively sampled 2D vertices (\ref{chamfer:vert_cloud}).}
  \label{fig:chamfer_loss}
\end{figure*}

Our model is described in Figure~\ref{fig:overview}, which shows how we build on the HMR approach. During inference, the HMID system only requires an RGB image sequence to extract the biometric parameters. During training, silhouettes (person foreground) and 2D keypoints are extracted and used to augment the biometric losses with additional losses that refine the HMR internal representation (see Figures~\ref{fig:chamfer_loss}, \ref{fig:pose_loss}). The silhouettes are computed using the Self Correction for Human Parsing (SCHP) model~\cite{li2020_schp}. The 2D keypoints are computed using the VIBE~\cite{kocabas2020_VIBE} model.

We train HMID on a subset of BRS~\cite{cornett2023expanding}. We split BRS into mutually exclusive training and validation sets. Each subject has two clothing sets and we compute metrics using a gallery split (indoor sequences) and a probe split (outdoor sequence, different clothing set compared to the probe). Each subject has several video sequences in different circumstances (controlled/field, random/structured movement, distances varying from close to 100m, 200m, etc). The augmentation is described in  \ref{sec:training:input}. 

\subsection{Input and Augmentation}
\label{sec:training:input}

The input during training consists of a sequence of images with corresponding silhouettes and 2D keypoints (Figure~\ref{fig:overview}). We select a random sub-sequence for each video in the training dataset with the following augmentations.

\begin{enumerate}
    \item Each image in the input sequence is resized randomly to $(s, s)$ where $s$ is sampled uniformly from $[48,224]$ before being upscaled to $(224,224)$ to simulate image quality loss.
    \item A consistent color jitter augmentation is also performed for the sub-sequence to simulate clothing color variation (in addition to actual clothing variation in the data).
\end{enumerate}

\subsection{Training losses}
\label{sec:training:losses}

We describe the training losses below.

\paragraph{Biometric Loss:} We use the arc-margin loss \cite{deng2022_arcface} using the mean biometric feature over the input sequence. Other losses we explored include the hard triplet loss as well as a combination of hard triplet and arc-margin. The hard triplet loss uses a tuple of (\emph{anchor, positive example, negative example}) where the anchor and positive examples have clothing variation which was slower and had poorer performance. 

\paragraph{Shape Loss:} We use the Chamfer distance between the silhouette cloud and the adaptively sampled SMPL model vertices projected onto the 2D image. The Chamfer loss is the sum, over each point in a silhouette, of the distance of the point to the point closest to it in the other silhouette (computed in both directions). The silhouette and vertices cloud are described in Figure~\ref{fig:chamfer_loss}. This loss tries to move the 3D SMPL model silhouette to match the estimated human body model silhouette. We adaptively sample the vertices so that the 2D image coordinates of the vertices image is more uniformly distributed. 

\paragraph{Pose Loss:} The 2D pose loss is obtained by the Mean Squared Error (MSE) between the 2D keypoint output of the model being trained and the pre-computed 2D keypoints. This error has the same units as the Chamfer distance (pixel error). The 2D keypoints are illustrated in Figure~\ref{fig:pose_loss}. This loss serves to anchor the important joints as well as maintain the pose estimates.

\paragraph{Shape consistency loss:} While the pose of the subject may change within a sequence  (i.e., the body is in motion), the subject shape should stay the same. This is enforced by adding a shape consistency loss to minimize the changes in the SMPL model shape parameters ($\beta$) within a sequence. This is enforced by minimizing the temporal variance of the shape parameter, $\sum_{i}\|\beta_i-\beta_\textrm{mean}\|$.

\begin{figure}[tb]
  \begin{subfigure}{0.49\linewidth}
    \centering
    \includegraphics[width=\linewidth]{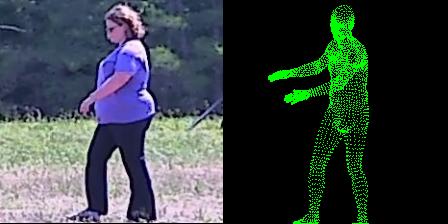} 
    \includegraphics[width=\linewidth]{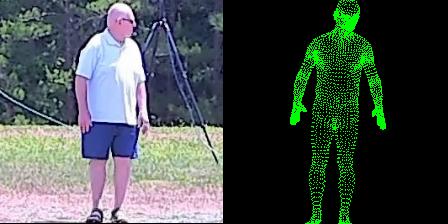}
    \includegraphics[width=\linewidth]{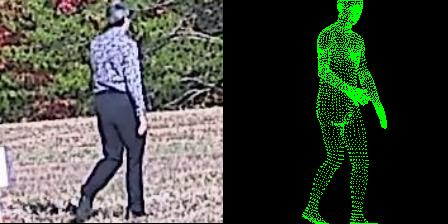}
    \includegraphics[width=\linewidth]{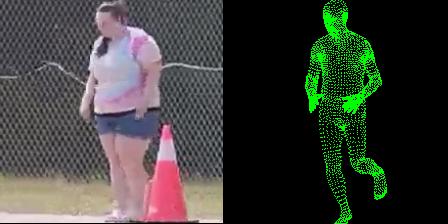}
      \caption{Pretrained HMR model}
  \end{subfigure}
  \hfill
  \begin{subfigure}{0.49\linewidth}
    \centering
    \includegraphics[width=\linewidth]{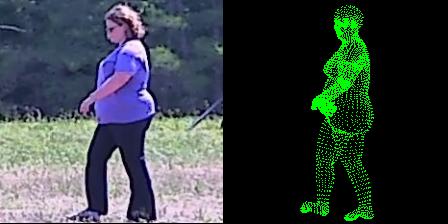} 
    \includegraphics[width=\linewidth]{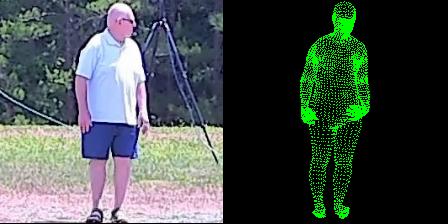}
    \includegraphics[width=\linewidth]{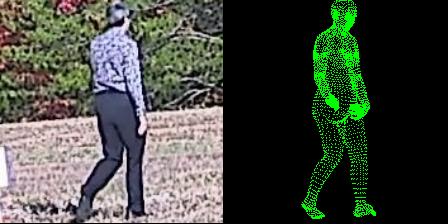}
    \includegraphics[width=\linewidth]{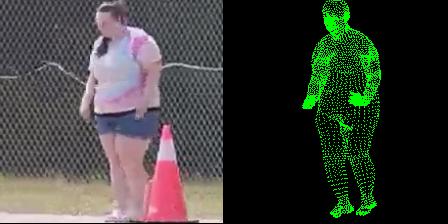}
    \caption{HMID model}
  \end{subfigure}
  \caption{SMPL human body model estimation before and after fine-tuning for shape. The images in column (a) show the input and output of the pre-trained HMR model. The images in column (b) show the output of our HMID model that has been trained with pose, shape and biometric loss. The body shape appears much closer to the actual shape of the subject in (b) although there is some distortion.}
  \label{fig:fine-tuning}
\end{figure}

The HMR model was pre-trained on the VIBE dataset, which has limited subject variation. We found it necessary to fine-tune our HMR-based model using the shape and pose losses on the more extensive BRIAR collection, in addition to using the biometric loss. The output of the pre-trained HMR from the VIBE model and the output of our HMR-based model that has been finetuned for shape is shown in Figure~\ref{fig:fine-tuning}. 

In \cite{li2020_end-to-end}, a shape loss using silhouettes is also applied to improve the shape information extracted by the HMR model but their loss is defined as a rendering error, following the work of~\cite{kato2018neural}: a differentiable renderer is added to the output of the HMR model that estimates the person's silhouette. The loss is then computed as the sum of the squared pixel-value errors between the estimated and actual silhouettes (where pixels are valued $1$ in the foreground and $0$ in the background). This pixel error then drives the image position of the projected mesh points and, in turn, the HMR shape parameters through back-propagation. Since the silhouette pixels are simply binary valued, the error gradients resulting from poor shape alignment are often a weak driver of the 3D shape parameters. Instead, we use a Chamfer loss~\cite{ravi2020accelerating}, which is a much closer and more explicit representation of point-to-point geometric alignment errors between the silhouettes. Since Chamfer loss relies on nearest-neighbor distances, it is only an approximation of the true shape error, but when shapes are very close (e.g., two human bodies with roughly the same pose), this approximation converges well.

\subsection{Inference}
\label{sec:training:inference}

During inference time, the model takes as input a sequence of RGB images. The sequence is split into sub-sequences of length 5 and each sub-sequence produces an ID feature vector and a median view estimate (roll, pitch and yaw) in addition to the other HMR outputs. We use the estimated view (Pitch and Yaw angles) to determine the bin in which the feature is aggregated.

\subsubsection{Representation and merging}
\label{sec:training:inference:representation}

\begin{figure*}[tb]
    \centerline{\includegraphics[width=\linewidth]{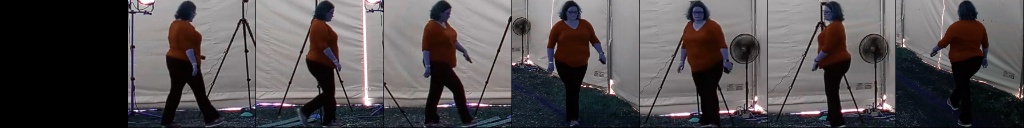}}
    \centerline{\includegraphics[width=\linewidth]{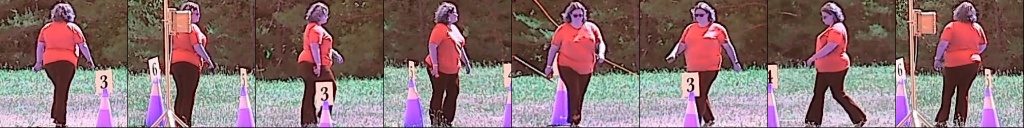}}
    \centerline{\includegraphics[width=\linewidth]{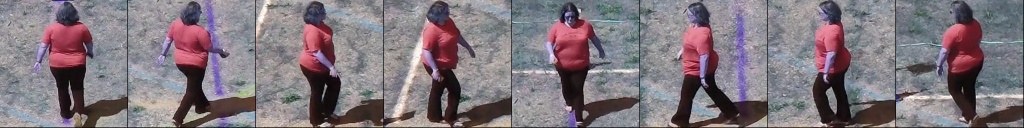}}
    \caption{Representative image for each view where the number of viewing bins is 8 based on yaw angle. The bins correspond to the intervals given by $[-\pi/8, \pi/8), [\pi/8,\pi/4), [\pi/4, 3\pi/8), \cdots$. The top row shows an example from a gallery sequence (controlled) and the other rows show corresponding probe (field) sequences where clothing and lighting differ. The third row is an example sequence shot from a UAV and represents a pitch angle of $\sim 30^\circ$. Not all view bins are occupied in the gallery or probe. }
    \label{fig:dossier}
\end{figure*}

We use a view-based representation scheme where the features are aggregated into bins based on the view (yaw and pitch angles). The pitch and the yaw angles are estimated by our HMID model (in addition to the biometric feature, shape and pose). We record a feature and an occupancy scalar (number of features in that bin) for each bin. We used either 1, 4 or 8 yaw bins (number of pitch bins=1) in our experiments. Features are aggregated in each bin using the \emph{mean}, \emph{median} or \emph{best}\footnote{The \emph{best} aggregation method computes inliers based on an adaptive threshold that is computed as the 90th percentile of distances between all pairs of features from a given sequence. The mean is then computed from the inlier features. This is intended to remove outlier features in the sequence} method. Our experiments suggests that mean aggregation is as good or slightly better than the median aggregation method. Each video produces a set of features (one feature per view bin). When a subject has multiple video sequences in the gallery, the feature sets are merged by using a weighted mean of the features (the features in a bin are weighted by their occupancies). A feature set with 8 yaw bins is illustrated in Figure~\ref{fig:dossier}.

\subsubsection{Matching} 
\label{sec:training:inference:matching}

Feature matching is straightforward when we use only one bin and is computed as the angular distance between the features. A lower distance implies a better match. When we use more than one bin, we compare features in the most occupied bin. E.g. to compare a feature set $F_1=\{f_1^{i}\}_{i=1}^{N}, \{o_1^{i}\}_{i=1}^{N}$ with $F_2=\{f_2^{i}\}_{i=1}^{N},\{o_2^{i}\}_{i=1}^{N}$, we select bin $i = \arg\max_i(o_1^{i}o_2^{i})$,
if there exists a bin in which both features have occupancy greater than 0. We then use the feature in that bin to compute the distance. If there is no bin with common occupancy, then we compute the distance by collapsing all bins (i.e. as if the number of bins=1). Our current work is focused on using an attention-based neural network to learn a general method of computing scores between two feature sets. 

\section{Experiments}
\label{sec:exp}

In our experiments, we trained our model on a subset of BRS~\cite{cornett2023expanding}.%
\footnote{The system was trained on a single GPU; either NVIDIA RTX A6000 (48GB memory) and NVIDIA GeForce RTX 3090 (24GB memory) were used in experiments.} 
We split BRS into a training split (2547 videos, 418 subjects) and validation validation split (172 videos, 86 subjects) with mutually exclusive subjects. Both training and validation sets are split into gallery (sampled from controlled data) and probe sets (sampled from field data) to compute metrics. We compute and report metrics on the validation set for ablation studies. 
In each case, we list the True Acceptance Rate (TAR) at False Acceptance Rate (FAR)=$1\%$ and the rank10 accuracy. We record the model checkpoint based on the highest rank10 accuracy and TAR@FAR=1\% scores on the validation set during training. 
We note that the model is \emph{not trained on any subject in the validation set} and it is therefore a reasonable estimate of the performance of the model on the test dataset. 
We also show results on an extended BRS validation dataset that includes pitch variation (\S\ref{sec:exp:pitch}), the actual BRIAR Test Set (BTS) (\S\ref{sec:exp:briar}) and a completely different USF HID dataset (\S\ref{sec:exp:usf}).

\subsection{Ablation studies}
\label{sec:exp:ablation}

In this section, we explore the effects of various losses, models and input parameters as well as matching algorithms\footnote{These and other variations result in a sequence of versions of HMID models indicated throughout the result tables using versions, \emph{i.e.} ``Ver\#''}. We test on the BRS validation split described above. The results are summarized in Table~\ref{tab:ablation}. 
\input{ablation}
\begin{enumerate}
    \item Training losses: We explore the effectiveness of various losses on the biometric accuracy. We note that in the absence of HMR losses, the biometric accuracy is much lower. We also note that arc-margin loss is more effective than hard triplet loss. It is possible to use both arc-margin and hard triplet loss, but there is no gain in performance. 
    \item Feature vectors: There are several options for selecting the feature to use for biometrics. One option is the \emph{raw} 2048D output of the core ResNet50 model in HMR. Another option is the 10D \emph{shape} component of the SMPL model ($\beta$) as used in \cite{li2020_end-to-end}. Yet another option is a \emph{fusion}, i.e. concatenation of the \emph{raw} and the \emph{fusion} vectors. In each case, a dense layer is added on top to output a 512D feature vector. We note that the raw feature performs the best, and the fusion version is slightly less accurate.
    \item Input parameters: In this experiment, we vary the length of the input sequence used in training and inference. Longer sequences have more information, but are expensive in terms of memory. We note that the performs seems to plateau for a sequence length of 5.
    \item Matching methods: The HMR model produces a feature vector for every input sequence. Given a video sequence, we typically get tens or hundreds of feature vectors as well as corresponding view angles. We can aggregate these feature vectors using the \emph{mean}, \emph{median} or \emph{best} methods (see \S\ref{sec:training:inference:representation}) into a single bin irrespective of view. We can also aggregate the features into different bins based on the viewing angle and perform matching as described in \ref{sec:training:inference:matching}. We show results for different number of yaw bins (1, 4 and 8). We note that the best performance is for \emph{mean} aggregation and using just a single view point. While using more bins can improve performance, we believe that we need a more sophisticated matching algorithm for comparing feature sets that do not have the same occupancy structure. 
\end{enumerate}

\subsection{Pitch variation}
\label{sec:exp:pitch}
We also study the effect of variation in the view pitch angle in the training and test data in Table~\ref{tab:pitch}. We introduce field examples into the (probe) dataset that have been captured from drones and have a pitch angle of $\sim 30^{\circ}$ as shown in Figure~\ref{fig:dossier}. We show that when trained without such examples  (HMID ver\#1) but tested on examples where there is significant pitch variation, the performance degrades. However, when examples with pitch variation are added to the training set (HMID ver\#15), the performance is better.

\subsection{BRIAR evaluation}
\label{sec:exp:briar}

We compare the models using the BRIAR~\cite{cornett2023expanding} 3.1.0 evaluation protocol in Table~\ref{tab:bts}. The numbers are reported for the case where the sequences from Gallery 1 and 2 are combined (a total of 541 subjects including confusers) and the different BRIAR probe sets were tested, including the combined probe set Probe ALL, with 6598 sequences. We compare our HMID models (ver\#15 trained with pitch variation and ver\#1 trained without) with our HMR model (same architecture as HMID but without shape and pose loss). We note that the HMID models shows a significant improvement in the performance over the HMR model that does not use HMR losses for shape and pose. We also note that including the pitch variation data in the training improves the performance on the challenging BRIAR evaluation dataset that includes long-range sequences as well as sequences from elevated view points.

HMID also compares well on BRIAR data against the system reported in~\cite{myers2023recognizing} (their best fusion model). Their system is evaluated using a 60-person gallery subset of BTS, resulting in a TAR at $1\%$ FAR of $25\%$,  a rank10 score of $88\%$ and rank20 score of $95\%$. Our BRS validation split test has 90 gallery subjects ($50\%$ more people). On this, HMID ver\#1 has a TAR at $1\%$ FAR of $~49.4\%$ and almost the same rank10 score of $87.4\%$. Our test of HMID ver\#15 on the full BRIAR Test Set of 530 gallery subjects (compared to 60) results in a TAR at $1\%$ FAR of $38.9\%$ and a rank20 of $62.1\%$.

\input{briar}

\subsection{USF evaluation, a cross-collection study}
\label{sec:exp:usf}

\begin{figure}[tbh]
    \centerline{\includegraphics[width=\linewidth]{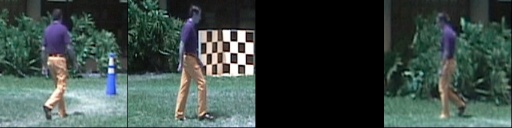}}
    \centerline{\includegraphics[width=\linewidth]{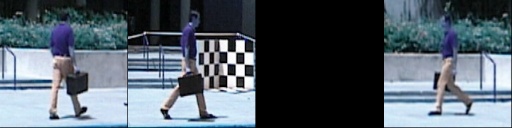}}
    \centerline{\includegraphics[width=\linewidth]{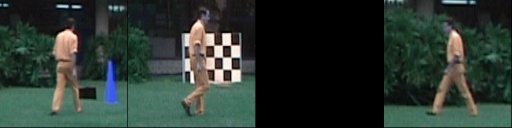}}
    \caption{Sample images from the USF dataset. }
    \label{fig:usf}
\end{figure}

We also compare the performance of our model (HMID ver\#1) on the USF dataset~\cite{sarkar2005_hid} in Table~\ref{tab:usf}. An example from the USF dataset is shown in Figure~\ref{fig:usf}. This is a cross-collection study: our model has been trained only on the BRIAR dataset and not fine-tuned on the USF dataset. Our model also does not explicitly use gait information as our approach is a general approach for any kind of pose or activity.  We note that our model out-performs the silhouette based approach in \cite{sarkar2005_hid} including the cases where the subject is carrying a briefcase. The two cases (K and L) where our model does not perform very well is when there is a time gap of 6 months between the probe and the gallery, though overall, it performs better than \cite{sarkar2005_hid} in those cases. 

\input{usf}

\section{Conclusion}
\label{sec:conclusion}

We have developed a model-based approach for biometric identification that does not require silhouettes. We train the model to estimate the human body model (including view, pose and shape) and base the biometric embedding on the intermediate feature. The advantage of this approach is that the model is robust to partial inclusions (such as briefcase) and does not rely on gait signatures or silhouettes. We show that the model trained on the BRIAR dataset outperforms the models developed on the USF dataset while evaluated on the USF dataset. Our results also show the promise of using model-based approaches that are robust to the challenges of biometric identification at range and elevation on the BRIAR evaluation dataset. 

Our approach can be extended by using more sophisticated human body models (such as STAR~\cite{osman2020_STAR} as compared to SMPL). We can also incorporate a recursive method for estimation of human body model parameters to improve temporal consistency in the model parameters. Our approach allows us to train on human body pose datasets as well as biometric datasets such as BRIAR. While our initial results show that aggregating features based on view does not improve biometric performance, we believe that attention-based methods for aggregating and matching probe and gallery sequences will improve the results further and is a work in progress.

Additional directions could include the study of a wider range of human activities and poses, and also an evaluation on data with changes in subject age.

%% file: ablation.tex
\begin{table*}[htb]
    \caption{Ablation study results on BRS validation split. The losses are HMR (shape, pose, and temporal shape consistency losses), ArcMargin and Hard Triplet. The biometric features used are \emph{raw, shape,} and \emph{fusion} (see \ref{sec:exp:ablation}.2).  The seq. length is the number of images in the input sequence. The matching algorithms include the aggregation method (mean, median, best) and the number of yaw bins used in the feature set (1, 4, 8). The True Accuracy Rate (TAR) is reported for a False Accuracy Rate (FAR)=$1\%$. The first row represents our default model; any cell with a ``-'' means that the default option is used for that system parameter. }
  \small
  \begin{center}
    \label{tab:ablation}
    \begin{tabular}{|l|r|c|c|c|c|c|c|c|r|r|} 
      \hline
      \hline
      \multicolumn{2}{|l|}{Experiments} & \multicolumn{3}{c|}{Losses} & {Model} & Input & \multicolumn{2}{c|}{Matching} & \multicolumn{2}{c|}{Metrics} \\
      \hline
   Variation& Ver\# &{HMR} & {ArcMargin} & {Triplet} & {Feature} & {Seq. len.} & Agg.   & Yaw bins & {Rank10} & {TAR}  \\  \hline 
    Default & 1 &Y     & Y           & N         & raw       & 5           & mean   & 1        &   \textbf{87.4}   & \textbf{49.4}   \\
                                                                                                                     \hline 
\multirow{3}{*}{Losses}                                                                                                                     
            & 2 &N     & Y           & N         &   -       & -           & -      & -        &   69.0   & 20.7   \\
            & 3 &Y     & N           & Y         &   -       & -           & -      & -        &   63.2   & 17.2   \\ 
            & 4 &Y     & Y           & Y         &   -       & -           & -      & -        &   88.5   & 42.5   \\ 
                                                                                                                       \hline 
\multirow{2}{*}{Input}                                                                                                                     
            & 5 &-     & -           & -         &   -       & 3           & -      & -        &   85.1   & 41.4   \\
            & 6 &-     & -           & -         &   -       & 9           & -      & -        &   83.9   & 40.2   \\
                                                                                                                       \hline 
\multirow{2}{*}{Model}                                                                                                                     
            & 7 &-     & -           & -         & shape     & -           & -      & -        &   34.5   & 3.45   \\ 
            & 8 &-     & -           & -         & fusion    & -           & -      & -        &   82.8   & 44.8   \\ 
                                                                                                                       \hline 
\multirow{4}{*}{Matching}                                                                                                                     
            & 9 &-     & -           & -         & -         & -           & median & -        &   \textbf{87.4}   & 48.3   \\ 
            &10 &-     & -           & -         & -         & -           & best   & -        &   79.4   & 44.8   \\ 
            &11  &-     & -           & -         & -         & -           & -      & 4        &   81.6   & 40.2   \\ 
            &12  &-     & -           & -         & -         & -           & -      & 8        &   80.5   & 44.8   \\ 
      \hline
      \hline
    \end{tabular}
  \end{center}
\end{table*}

\begin{table}[htb]
  \caption{We compare the performance of the model when trained and tested on an extended BRIAR validation dataset that has sequences with view pitch angle $\sim 30^\circ$. The number of gallery and probe sequences in the original and extended test dataset is also listed in the table. }
  \small
  \begin{center}
    \label{tab:pitch}
    \begin{tabular}{|r|l|l|r|r|r|r|} 
      \hline
      \hline
      \multicolumn{3}{|c|}{Pitch Variation} & \multicolumn{2}{c|}{Size} & \multicolumn{2}{c|}{Metric} \\ \hline
      Ver\# & Train  & Test    & Gallery & Probe & Rank10        & TAR       \\  \hline 
      1  & N      & N       & 86      &  86   &        {87.4} &        {49.4} \\
      15 & Y      & N       & 86      &  86   &        {90.8} &        {49.4} \\
      1  & N      & Y       & 91      & 182   &        {85.7} &        {43.4} \\
      15 & Y      & Y       & 91      & 182   & \textbf{92.3} & \textbf{51.1} \\
      \hline
      \hline
    \end{tabular}
  \end{center}
\end{table}

%% file: briar.tex
\begin{table}[htb]
  \caption{Performance on the BRIAR Test Set (BRIAR evaluation protocol 3.1.0). The probes are FIC (Face Included Control), FRC (Face Restricted Control), FIT (Face Included Treatment) and FRT (Face Restricted Treatment) and ALL. The gallery is the combined Gallery0 and Gallery1 sets and includes 541 subjects. The Rank20 and TAR at FAR=$1\%$ is shown for the three models selected from Tables~\ref{tab:ablation} and \ref{tab:pitch}.}
  \small
  \begin{center}
    \label{tab:bts}
    \begin{tabular}{|l|r|r|r|r|r|r|} 
      \hline
      \hline
      \multicolumn{1}{|c|}{Probe} & \multicolumn{2}{c|}{HMR Ver\#2} & \multicolumn{2}{c|}{HMID Ver\#1} & \multicolumn{2}{c|}{HMID Ver\#15} \\ \hline
      Name (\#)     &         Rank  &         TAR   &         Rank  &         TAR   &         Rank  &         TAR   \\  \hline 
      FIC  (1259)   &        {37.8} &        {18.7} &        {56.9} &        {32.6} & \textbf{70.6} & \textbf{50.2} \\
      FRC  (353)    &        {38.8} &        {13.9} &        {53.8} &        {33.1} & \textbf{65.7} & \textbf{50.4} \\
      FIT  (1596)   &        {34.1} &        {14.9} &        {45.2} &        {24.8} & \textbf{59.0} & \textbf{33.7} \\
      FRT  (3390)   &        {36.6} &        {15.3} &        {48.0} &        {26.3} & \textbf{60.0} & \textbf{36.2} \\ \hline
      All  (6598)   &        {36.3} &        {15.3} &        {49.3} &        {27.4} & \textbf{62.1} & \textbf{38.9} \\
      \hline
      \hline
    \end{tabular}
  \end{center}
\end{table}

%% file: usf.tex
\begin{table}[ht]
  \caption{Model performance on USF dataset for different probes (HMID ver\#1). The gallery sequence was on Grass, shoe type A and the right camera without briefcase and has 122 subjects. The number of subjects in each probe varies from 33 to 122 and is listed in the second column. The TAR was computed at FAR=$1\%$ and the descriptions of the various probes are in \cite{sarkar2005_hid}}
  \small
  \begin{center}
    \label{tab:usf}
    \begin{tabular}{|c|r|r|r|r|r|r|r|} 
      \hline
      \hline
      \multirow{2}{*}{Pr.} & \multirow{2}{*}{Num.} & \multicolumn{2}{c|}{Rank1} & \multicolumn{2}{c|}{Rank5} & \multicolumn{2}{c|}{TAR}\\ \cline{3-8}
                             &     & USF & Ours  & USF & Ours  & USF & Ours   \\\hline
A    & 122 &  73 & \textbf{100.0}  &  88 &  \textbf{100.0}  & 52  &  \textbf{100.}  \\
B    &  54 &  78 & \textbf{88.8}  &  \textbf{93} &  92.5 & 48  &  \textbf{88.8}  \\
C    &  54 &  48 & \textbf{88.8}  &  78 &  \textbf{92.5} & 32  &  \textbf{88.8}  \\
D    & 121 &  32 & \textbf{77.3}  &  66 &  \textbf{97.3} & 24  &  \textbf{64.3}  \\
E    &  60 &  22 & \textbf{66.6}  &  55 &  \textbf{83.3} & 16  &  \textbf{57.4}  \\
F    & 121 &  17 & \textbf{77.3}  &  42 &  \textbf{97.3} & 10  &  \textbf{64.3}  \\
G    &  60 &  17 & \textbf{66.6}  &  38 &  \textbf{83.3} & 12  &  \textbf{57.4}  \\
H    & 120 &  61 & \textbf{96.5}  &  85 &  \textbf{100.0} & 46  &  \textbf{98.2}  \\
I    &  60 &  57 & \textbf{84.4}  &  78 &  \textbf{91.3} & 48  &  \textbf{84.4}  \\
J    & 120 &  36 & \textbf{96.5}  &  62 &  \textbf{100.0} & 22  &  \textbf{98.2}  \\
K    &  33 &  \textbf{3} & \textbf{3.0} &  \textbf{12} &  9.0 & 0   &   \textbf{4.0}  \\
L    &  33 &  3  &  \textbf{8.1}  &  15 &  \textbf{19.3} & 0   &   \textbf{9.2}  \\ \hline
      \hline
    \end{tabular}
  \end{center}
\end{table}
%